# Vision System for AGI: Problems and Directions

Alexey Potapov[1,2], Sergey Rodionov[1,3], Maxim Peterson[1,2], Oleg Scherbakov[1,2],
Innokentii Zhdanov[1,2], Nikolai Skorobogatko[1,3]

[1] SingularityNET Foundation, The Netherlands
[2] ITMO University, St. Petersburg, Russia
[3] Novamente LLC, USA
{pas.aicv, astroseger, maxim.peterson}@gmail.com,
{avenger15, scherbakovolegdk}@yandex.ru, nicksk@mail.ru

**Abstract.** What frameworks and architectures are necessary to create a vision system for AGI? In this paper, we propose a formal model that states the task of perception within AGI. We show the role of discriminative and generative models in achieving efficient and general solution of this task, thus specifying the task in more detail. We discuss some existing generative and discriminative models and demonstrate their insufficiency for our purposes. Finally, we discuss some architectural dilemmas and open questions.

## 1 Introduction

Within "Good Old-Fashioned Artificial Intelligence", vision was considered as a peripheral function, which doesn't have a direct relation to the mind, which was more associated with the knowledge-based symbolic reasoning. This situation has been preserved for many cognitive architectures (CAs), not only purely symbolic, but also hybrid, which cannot process images by themselves and require additional modules. However, such external perception modules appear to be not tightly integrated into CAs limiting these CAs in their ability to interact with the physical world.

At the same time, the idea that the basis of natural intelligence consists in pattern recognition is quite usual. In turn, experts in computer vision sometimes joke that thinking is just the upper level of the visual system. However computer vision has for a long time being developed in a relative isolation from the AI field, while the purely emergent CAs, going from perception upwards, are far from solving symbolic tasks that are the prerogative of the human mind.

Moreover, such isolated tasks as object detection and recognition, motion analysis, stereovision, shape from shading (or even more narrow tasks, e.g. shadow detection) are studied in computer vision. The problem of how a general vision system should be built is mostly not addressed, and we believe it cannot be reduced to a set of narrow tasks. This problem is especially relevant in the field of AGI.

In this paper, we consider the question how to proceed towards the goal of creation of a sufficient vision system for AGI. Here, we do not propose to limit ourselves to considering only those technical solutions that can work in real time on modern computers, however, consideration of resource constraints is fundamentally necessary for developing a potentially realizable vision system. For example, the unified model of perception and action selection given within AIXI [1] is not sufficient since even its rough approximation [2] can deal only with low-dimensional inputs.

## 2  Vision Task

*Separation of the vision subsystem*
The ultimate task of intelligence consists in calculating probability $P(a_t|x_{1:t},a_{1:t-1})$ of taking action $a_t$ at time $t$ ($x_t$ unite current sensory and reinforcement signals) in order to maximize expected future rewards. AIXI uses universal (Solomonoff) induction [3], which predicts future observations via marginalization over all computable generative models consistent with the agent-environment interaction history. Consider the case of pure induction:

$$P_U(x) = \sum_{\mu:U[\mu]=x*} 2^{-l(\mu)},$$

where $\mu$ is a program for universal machine $U$, generating a string with prefix $x$. The probability distribution over continuations $x'$ of $x$ is $P_U(x'|x)=P_U(xx')/P_U(x)$.

Of course, enumeration of all models at each time step is computationally infeasible. Consistent models should be somehow "cached", which goes far beyond sensory system and makes up half of the intelligence (the second half is decision making). Is it possible to draw a boundary between the sensory subsystem and the rest of intelligence (in particular, memory), or do we need a holistic model like AIXI?

The task of perception can be seen in the processing of the current data $x_t$ (or $x_{t-k:t}$ for small $k$), while the whole history should be dealt with by the memory.

Without loss of generality, one can consider environment models (not necessarily Markovian) with internal states $z_t$. Then, let $\mu(z_t|z_{t-1},a_t)$ be an environment model and $o(x_t|z_t)$ is an observation model (computable probability distribution). Indeed, one can assume $z_t=x_{1:t}a_{1:t}$ with trivial o, which, of course, doesn't give any advantage. However, one can hope that $z$ can represent the interaction history much more compactly. Then, the task of perception is to infer $z_t$ from $x_t$ with the use of priors $\mu(z_t|z_{t-1},a_t)$.

This task is not simpler than universal induction since to predict $x_t$ one still needs to marginalize over all possible models, and for each model to calculate the probability of the interaction history (marginalizing over all $z_{1:t}$).

However, we can (or should for the sake of efficiency) approximately solve the task of perception in assumption of the fixed $\mu$ and o. Then, this task will consist in inferring posterior probabilities over $z_t$ given $x_t$:

$$P(z_t|x_t) = \frac{P(x_t|z_t)P(z_t)}{P(x_t)} = \frac{o(x_t|z_t)\int \mu(z_t|z_{t-1},a_t)P(z_{t-1}|x_{t-1})dz_{t-1}}{\int P(x_t|z_t)P(z_t)dz_t}, \qquad (1)$$

if an AGI system maintains an uncertain state representation in the form of $P(z_t|x_t)$. Simplification $P(z_t)=\mu(z_t|z_{t-1},a_t)$ can useful for analysis, but imprecise.

Thus, the task of perception can be reduced to (1), although the task of learning the models should also be accounted for.

*Discriminative and generative models*

Does AGI really need to reconstruct a generative model of the environment? Indeed, the main approach in reinforcement learning (RL) is model-free (although the model-based approach is advocated as a more adequate, e.g. [12]). Although it assumes some class of environments, their models are not explicitly reconstructed. Instead, value functions are estimated or policies are directly learnt, which can be treated as discriminative models since they describe conditional instead of joint probabilities.

Discriminative models are intensively used in computer vision and machine learning also. Traditional and most successful (at least, in pattern recognition) deep learning models are discriminative. However, the shortcomings of these models have also been generally recognized recently.

In particular, discriminative models don't support unsupervised, semi-supervised, or one-shot learning. Transfer learning with these models is also difficult. For example, in the case of reinforcement learning, the policy or value function should be retrained from scratch even for the same environment, but modified reward function [4]. In terms of AGI, we can say that discriminative models belong to narrow AI.

Generative models possess the required flexibility and support all the mentioned forms of learning because they "explain" data, but not just predict target variables. The possibility to generate data is not usually an aim, but means to ensure that the description of data is complete, and no information is lost, thus, enabling criteria for any kind of learning. Discriminative models throw away information, which is irrelevant to target variables, and we don't know its amount, thus, learning criteria based on the prediction of target variables are to be used.

In the context of AGI, we cannot limit ourselves to the consideration of discriminative models, regardless of whether they map observations to actions to or labels for detected objects. <u>Ultimately, it is necessary to state the task of vision as a task of reconstruction of a latent description of a scene within a trainable generative model</u>.

Unfortunately, inference over generative models is computationally demanding not only in universal induction, but also in more specific cases (inference can be inefficient even in a very limited case of graphical models), so they are also not sufficient.

We consider a discriminative model as a result of specialization of a general inference procedure in projection onto a certain generative model [5]. So, the properties of models of both types are understandable. In generative models, the inference process is separated from the model itself, and the models can be flexibly changed. Discriminative models can be viewed as efficient, but narrow inference procedures over certain generative models. Any changes to the generative model (albeit not presented explicitly) will affect this specialized inference procedure in a non-trivial way. Discriminative models are like reflexes, which are developed to solve narrow tasks with maximum efficiency, but which are badly applicable outside their scope.

But shouldn't a developed vision system processing huge volumes of information normally act as a specialized discriminative model which maps observations $x_t$ into their latent code $z_t$? Seemingly, it should. Of course, AGI should be able to learn to recognize new classes of objects. However, natural images contain typical regularities, for the extraction of which a discriminative model can be trained once.

The practice of using deep convolutional neural networks (DCNNs) shows that networks trained discriminatively on large databases such as ImageNet can be successfully used when solving more specialized recognition tasks. However, in order to get really good results, fine-tuning of pre-trained networks is required. This process affects the upper level features, while the lowest levels remain mostly unchanged.

In the human visual system, there are also specialized discriminative models, e.g., for the face analysis. It is difficult to say whether new discriminative models are formed for the analysis of specific images (e.g. tomograms), but even if this does not happen, and analysis of such images by our visual system remains suboptimal, this restriction is not necessary to reproduce in AGI. Thus, the higher the level in the discriminative vision subsystem, the more extendable it should be.

If there is no need to modify lower levels of the discriminative model, should the generative image model go down to the pixel level? Humans see images of scenes. These are not the images registered by the retina, but the reconstructed images. Indeed, we see not the pixel brightness, but the estimates of the reflective characteristics of the points on physical surfaces (as demonstrated by a number of visual "illusions").

If the generative model ended its work at the level of some convolutional features, then we would not see, e.g., hot pixels on monitors. On the other hand, when responding to sudden and rapid events, humans can perform adequate actions before understanding what they react to. This is similar to a quick inference by discriminative models processing images from pixels to scene descriptions. <u>Thus, both generative and discriminative models work at all levels of perception.</u>

However, it should be noted that the discriminative model $Q(z|x)$ is constructed as a variational approximation to the posterior distribution specified by the generative model with *fixed priors* $P(z)$: $P(z|x)=P(x|z)P(z)/P(x)$. But in accordance with (1), probabilities $P(z_t)$ are calculated at each time step using predictions $\mu(z_t|z_{t-1},a_t)$. Indeed, human vision system intensively uses such predictions to compensate for eye and body movements, to recognize objects in known dark rooms, etc.

Thus, the bottom-up processing of images by the discriminative model can only produce hypotheses with high likelihood (i.e. efficiently sample $z_t$ with high $o(x_t|z_t)$), while the generative model can propagate prior expectations $\mu(z_t|z_{t-1},a_t)$ top-down. Possibly, adaptive resonance [6] unites these processes into one procedure of iterative search for optimal $z_t$ accounting both for the likelihood and expectations.

Thus, <u>although the task of vision consists in inferring a latent description within the trainable generative model, a solution of this problem requires the construction of a system of discriminative models, both general purpose and specialized, with possible interaction with the generative model</u>.

*Learning*

A generative model $o(x_t|z_t)$ samples an image of the scene from its description. This description is multi-level – it includes both a list of objects, and information about visible surfaces, their reflectivity maps, reconstructed sources of illumination, etc. This can be treated as a 3D rendering engine with a library of objects, textures, etc.

Obviously, the models of new objects must be constantly learned. More difficult is the question regarding the scene rendering model. Should the AGI visual system learn the laws of light propagation, reflection, dispersion and refraction from scratch? The dimensionality of our world and its geometric laws? Indeed, a general intelligence would have to be able to reconstruct the appropriate environment model for any type of sensor on its own. On the other hand, there is no reason why we should not alleviate this problem for AGI systems from practical considerations by explicitly laying down or pre-training the inevitably necessary elements in the generative model (for example, the 3D representation of scenes and the laws of their projections).

However, our world is too diverse to take into account all the possible aspects of image formation, especially since AGI will need to perceive images of arbitrary objects (from atoms to galaxies) formed by special optical devices. In particular, although modern rendering engines can generate photorealistic images, they are not capable of generating any images that can be found in reality.

Thus, the generative image model $o(x_t|z_t)$ should be trainable, but the degree of this trainability and the content of priors are the questions for deeper discussion.

Apparently, the environment model µ should be learned mostly from scratch (although some general priors are necessary). This problem belongs to the field of AGI as such, and goes far beyond the scope of this paper, but we want to emphasize that arbitrary changes in µ can cause arbitrary changes in the space of latent states $z_t$ that render $o(x_t|z_t)$ obsolete. From the point of view of the sensor subsystem design, this space should be common to all possible environment models, and should be expandable, but not replaceable. Indeed, the acquaintance with the matter atomic structure does not force us to rewrite the entire content of our memory in new terms or retrain our vision system to account for the Maxwell's equations.

The discriminative vision subsystem must also be trainable at least to recognize new objects, but it might be necessary to retrain lower levels too (one can imagine an AGI system that has never before fallen into a snowfall and whose discriminative vision subsystem is not pre-trained on images obtained under such conditions).

Won't such learning disrupt the descriptions of previously recognized objects? To avoid this, the embedding space of the older objects should remain unchanged. Fortunately, this problem is solvable with the use of the generative model: descriptions formed by the discriminative model, should not just be useful for recognition or decision-making, but they should allow the generative model to reconstruct initial images.

*Memory*
We live in a very large environment, and instant observations $x_t$ contain not too much information about its partial state $z_t$ known to an agent. Thus, density $o(x_t|z_t)$ as a function of $z_t$ given $x_t$ will be very wide, and it is useless to require the discriminative subsystem to estimate it. It is natural to make estimations only for those hidden variables, information about which is contained in $x_t$ (i.e. the content of the current scene).

However, the estimation of probabilities even of relevant hidden variables is problematic for the purely discriminative model. In particular, we expect $z_t$ to contain some form of 3D reconstruction of a current scene. In practice, the task of simultaneous localization and mapping (SLAM) is considered, in which 3D coordinates of image points are not estimated by bottom-up image processing solely, but by matching these points with the earlier reconstructed map, that results in the map update also. Thus, the estimation of the visible part of $z_t$ should intensively use $z_{t-1}$, i.e. both read and update the memory. The solution of this task greatly depends on the memory organization (how the map is represented in it, to what extent this representation is trainable, and so on), that goes far beyond the vision problem only.

Nevertheless, the processing of stand-alone still images should also be supported (humans can perceive photos). Thus, some part of development of the vision system can be carried out in isolation from the rest of the AGI system. However, we should expect such vision system to learn not 3D scene reconstruction and separate objects, but only lower-level texture and contour-based 2D segmentation.

The vision is connected in a non-trivial way with the semantic vision also, which contains information about objects and their relations. The part-whole and is-a relations are used by the vision system and are partially formed with its help. The scene description can go down to the finest pixel-size details like specks of dust hanging in the air or grains of sand composing a texture.

Apparently, existing discriminative models are not that detailed. If they recognize a face, then not as a hierarchy of objects starting with individual hairs and wrinkles, but via features integrally describing square fragments of growing sizes regularly covering the image area (which makes it difficult to assign some semantics to such features even in the context of recognized larger objects).

Although specialized discriminative models can be trained for recognizing small objects of high importance, it seems that generative models should typically participate in the construction of the detailed scene description. These are the generative models, which "know" the structure of objects, and try to fit the parameters of this structure to observations. Indeed, humans usually are not conscious of all scene details, if they don't pay special attention to them (i.e. inference over generative models is controlled by general cognitive functions).

In fact, the generative model of images also does not need to operate with concepts such as each individual hair on the head or a speck of dust. Informationally, they are of little importance, and the generative model can consider their deviation from the background as noise. In general, generative and discriminative models can share all levels of the representation, so that the generative model will "draw" the image using "brushes" – the transposed filter kernels of the discriminative model.

Thus, in general, one should not expect to extract too many semantic categories as a result of the analysis of individual images. At least, video sequences with a varying point of view and with the (dis)appearance of objects should be analyzed in order to separate the concept of objects from their immediate sensory image. In addition, some usual object classes may be due to other sensory modalities or pragmatic criteria (it can even be argued that most categories are separated based on the ability to manipulate the relevant objects; for example, the notion of chair or cup is determined not so much by their visual features, but by their usage). Nevertheless, the trained visual system in an autonomous mode should be able to extract significant visual categories. We just should not demand the extraction of exactly the categories that we use. In general, however, it should be borne in mind that the boundary between the generative model of the environment for which the entire intelligence is responsible and the generative model of images turns out to be rather vague.

## 3 Frameworks

*Discriminative models*
DCNNs show outstanding results in image recognition. Are these networks sufficient to implement discriminative models for AGI vision?

First of all, the task of vision for AGI is much broader. What is needed is not just to recognize an object by its image, but to construct a scene description including all the objects with their shapes, poses, reflectance maps, etc. DCNNs are used also for the object detection, 3D reconstruction and semantic segmentation, but the results are not so great here. Neural solutions for the SLAM problem are also being developed (e.g. [7]), but their architectures are far from purely discriminative DCNNs.

Thus, either discriminative DCNNs should be used to construct some intermediate representation, on the top of which some other models (neural or not) will be built, or the whole formalism should be modified. "Object-oriented deep learning" [8] can be mentioned as one example of such modification. While it is difficult to say how effective can such extensions be, they look attractive from the point of view of uniting discriminative models with symbolic generative models and cognitive architectures.

Secondly, as noted, applied DCNNs are trained on the labeled data. Of course, pre-trained models can be used to build AGI, but training of new models will inevitably be required. In this case, manually labeled data will be absent, and the discriminative model will be taught on the basis of or in conjunction with the generative model (that is, the learning signal will be weaker than for discriminative training). In this connection, the efficiency of generalization will be critical, which is very low in classical DCNNs: they are not capable of generalizing beyond the area of the training sample, but only interpolate inside it.

For example, if a DCNN did not see objects of some class rotated in a certain range of angles, then it will not be able to recognize this class for new angles, even if it learned to recognize other classes for all angles. Does the discriminative model have to be rotation invariant in the same way as the invariance to shifts is ensured by using convolution? It is possible that with proper implementation this could be practical, but this does not solve the problem of weak generalization for arbitrary transformations. The source of this problem is a fixed system of links within the network so that a fragment of the network that implements some function is always applied to data coming from the same addresses (neurons). Convolutional networks go beyond such tight connectivity, but they apply the same network to different addresses in a fixed manner.

This problem can be approach in different ways, for example, by introducing dynamic addressing, as is done in models with external memory [9] or in capsule networks [10]. Unfortunately, capsule networks only partially solve the problem of weak generalization.

For example, we conducted the following experiment. We took eight MNIST digits (excluding 6 and 9), and trained CapsNets on six digits rotated arbitrarily and two digits (3 and 4) rotated in range [–45°, 45°]. The precision on the training set composed of 3 and 4 rotated within [–45°, 45°] was 99.04% for dynamic routing [10] and 98.27% for EM-routing [11], while for 180°±45° it appeared to be 1.05% and 12.92% correspondingly. The baseline DCNN model also showed 1.02% precision meaning that it systematically confuses rotated 3 and 4 with other digits, while EM-routing CapsNets recognize them on a level of random guess.

Thus, more powerful models are still to be developed. Alternatively, the responsibility for achieving the invariance to arbitrary transformations can be shifted to generative models and corresponding inference mechanisms.

For example, if the generative model learned to rotate arbitrary scenes, then the rotation angle can act as an unknown latent variable, which is not estimated by the discriminative model. The generative model (e.g. using EM-algorithm) can try to guess such angle, which lead to a self-consistent solution: the discriminative model produces such description of the generated rotated scene, which allows the generative model to render this scene (and the generative model can render the original image from the same description, but another value of the rotation). This can be considered as the mental rotation used by humans to recognize objects in unfamiliar perspectives.

Thus, discriminative models should not be necessarily capable of automatic learning of invariants, but the question whether to expand the existing formal neural models or not remains open. In turn, the need to extend discriminative architectures, both for solving wider problems, and for interacting with the generative model, is obvious.

*Generative models*

If in the case of discriminative models one can reconcile with their insufficient universality in favor of efficiency, then the requirements for the expressiveness of generative models are much higher. In particular, the generative model should be able (to learn) to generate images of the same object viewed from different angles.

Consider the following experiment. Let us take a deep convolutional adversarial autoencoder that receives a non-rotated image from the input, and the result of the reconstruction is compared with the rotated image, while the correct rotation angle is supplied as an additional latent feature (sine and cosine of an angle as two neurons). We teach the autoencoder to rotate digits from MNIST to all angles, while restricting the range of angles for 4 and 9. Figure 1 shows the results of reconstruction.

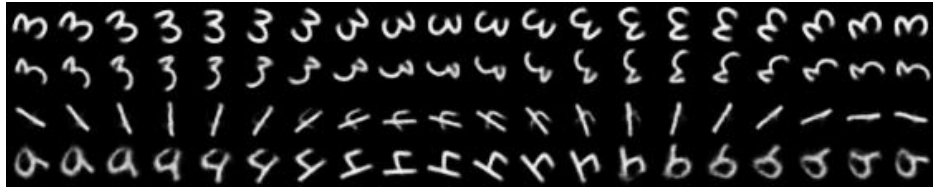
**Fig. 1.** Reconstruction results for different rotation angles (deep convolutional AAE)

As it can be seen, the autoencoder successfully reconstructs new images of digits rotating them at known angle (all images of 3 and central part for 4), but it transforms 4 to other digits trying to rotate it to a new angle.

The absolutely same effect takes place even if we try to train the autoencoder to reconstruct the shifted images. It might seem that convolutional networks should naturally perform the necessary generalization. Indeed, the transposed convolutional network can easily reconstruct the digit at any given location. However, the corresponding pattern in the highest level convolutional feature maps should somehow be activated. This is what dense layers going from the latent code cannot do. Connections going to different places in the feature maps are trained independently, so the network cannot transfer its experience in drawing 4 at one place to draw it at another place, and it will simply draw the digit with the most similar latent code which it knows how to draw at this specific place.

Thus, traditional generative neural networks cannot generalize spatial transformations independent of their content. Again, the question arises if we should extend the existing formalisms, and if yes, how specialized for vision should this extension be? For example, a specialized architecture can be crafted, which has a network for learning spatial transformations $(x',y')=f(x,y|\mathbf{w})$, which are then directly applied to images or feature maps. Similarly, one can train a network to transform 3D points. Apparently, this solution will be narrow, although it might be useful to the pragmatic AGI. But for us, more general solutions are more interesting.

For example, if we add second-order control neurons, which accept transformation parameters as input and influence the connection weights going from the latent code neurons to the highest level feature maps of convolutional autoencoder, then such network can learn to reconstruct arbitrary images independent on their content. We trained such network on the same data as the autoencoder. Figure 2 shows the results of reconstruction of both of the previously seen digit with new rotation angles, and reconstruction of a new symbol rotated on arbitrary angles. The network has only some problems with image corners, since they were always black in the training set, so it couldn't learn a mapping for pixels in them. Second-order networks are a general extension of ANNs, but the specific architecture we used here is rather specialized. Other solutions to the problem can be proposed, but their efficiency and generality for the AGI vision are the topics for further investigations.

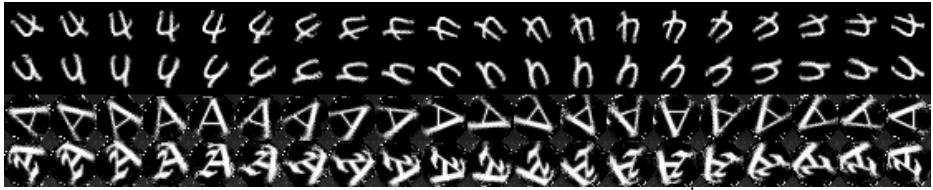
**Fig. 2.** Reconstruction results for different rotation angles ($2^{nd}$-order autoencoder)

Instead of extending neural networks one can try to use more general and expressive frameworks for specifying generative models like probabilistic programming. It is relatively easy to write a probabilistic program serving as a scene generative model. Unfortunately, the training of such models is computationally problematic, so one has to impose very strong priors in practice making the solution not too general.

Also, while discriminative models by themselves should not necessarily be capable of learning clear whole-part relationships, the latent description of a scene within the generative model should be ultimately expressed in terms of objects and their parts, thus, corresponding relations should be somehow learned. Existing frameworks neither for generative nor for discriminative models are powerful enough to do this.

## 4 Conclusion

The vision system should ultimately construct the scene description and participates in the reconstruction of the environment model. The generative subsystem guides the (unsupervised, one-shot, transfer) learning process and accounts for expectations, while the discriminative subsystem makes perception efficient in typical situations. However, many details are unclear.

– Should we require the capabilities to learn invariants and extract hierarchical relations from the discriminative models? If no (which seems biologically plausible), they should work with tight integration with the generative models, without which they will be useful only for forming reflective responses to stereotypical stimuli. But this doesn't mean that more powerful discriminative models cannot be developed.

– Should generative models be normally involved in image analysis? It seems, yes: purely discriminative models can be trained to solve not too narrow vision tasks on super-human level, but the necessity to propagate expectations is rather common. However, architectures with more emphasis on discriminative models are possible.

– Should the generative subsystem infer such latent variables, which are not estimated by the discriminative subsystem? It seems, yes, but only occasionally (i.e. to imagine a rotated object or a person wearing glasses, when recognition fails), since general inference over generative models is computationally demanding.

– How strong priors should be? This question is really controversial.

If we go down on the level of formalisms and implementations, the number of vague questions will increase. Should the generative and discriminative models be aligned on all levels? Should we use traditional neural networks for discriminative models? How should we extend existing formalisms for generative models? What are acceptable architectures for vision tasks beyond object recognition? And so on.

In general, we can conclude that existing frameworks and models are far from enough for implementing the vision system for AGI, especially in the generative part, which should be capable of rendering the images of scenes with new combination of objects in new poses. Also, tight integration of generative and discriminative models for efficient inference should be studied. This enables the consideration of the vision system as a part of AGI systems.